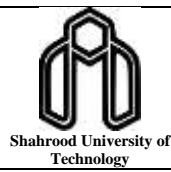

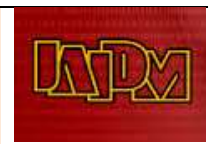

**Shahrood University of Technology**

**Research paper**

# Enhancing Aspect-based Sentiment Analysis with ParsBERT in Persian Language

Farid Ariai[1], Maryam Tayefeh Mahmoudi[2]*, and Ali Moeini[3]

*1. Department of Computer Engineering, University of Tehran, Kish International Campus, Kish, Iran.*
*2. Data Analysis & Processing Research Group, IT Research Faculty, ICT Research Institute, Tehran, Iran.*
*3. School of Engineering Science, College of Engineering, University of Tehran, Tehran, Iran.*

| Article Info | Abstract |
|---|---|

In the era of pervasive internet use and the dominance of social networks, researchers face significant challenges in Persian text mining including the scarcity of adequate datasets in Persian and the inefficiency of existing language models. This paper specifically tackles these challenges, aiming to amplify the efficiency of language models tailored to the Persian language. Focusing on enhancing the effectiveness of sentiment analysis, our approach employs an aspect-based methodology utilizing the ParsBERT model, augmented with a relevant lexicon. The study centers on sentiment analysis of user opinions extracted from the Persian website 'Digikala.' The experimental results not only highlight the proposed method's superior semantic capabilities but also showcase its efficiency gains with an accuracy of 88.2% and an F1 score of 61.7. The importance of enhancing language models in this context lies in their pivotal role in extracting nuanced sentiments from user-generated content, ultimately advancing the field of sentiment analysis in Persian text mining by increasing efficiency and accuracy.

## 1. Introduction

The expansion of the virtual space and social networks has led to the extraction, parsing, and analysis of a vast amount of structured and unstructured information. These data can be analyzed for various purposes using data mining, text mining, and web mining algorithms. Blogs, popular websites, and social networks serve as valuable sources of people's opinions and sentiments on various societal issues, which can be utilized for commercial, political, and social purposes. Nowadays, customers express their thoughts and emotions more than ever before, and sentiment analysis has become an essential tool for monitoring and understanding these sentiments. Automated customer feedback analysis, such as survey responses and social media comments, allows brands to understand what makes customers happy or dissatisfied, tailor their products and services to their needs, and measure customer satisfaction through pricing and customer service evaluations. With the proliferation of social networks, valuable tools for sharing opinions with others have emerged. In general, it is possible for a machine to find websites and opinions and summarize them, so we need an automated system to find relevant documents, structure the data, and generate summaries. Researchers have conducted extensive research in opinion mining and sentiment analysis in recent years. Sentiment analysis is an area that focuses on examining emotions, attitudes, opinions, and expressed states about various topics in texts using computational methods. The focus of sentiment analysis is on recognizing the emotions in texts or content, while sentiment classification focuses on detecting polarity. Both approaches utilize machine learning and natural language processing (NLP) to extract knowledge and information from opinions and texts. Opinion analysis is used to understand collective behavior regarding customer satisfaction, political movements, predicting the stock market, and in fields like psychology and medicine. In the context of customer satisfaction, people's opinions about the quality and pricing of e-commerce websites such as Digikala [1] are of great importance, and social networks provide a low-cost platform for collecting and analyzing such information and opinions. Various methods such as NLP, data mining or their combination are employed for opinion classification. Recent research has mostly focused on hybrid methods and pre-defined resources like WordNet or dictionaries. In sentiment analysis, most studies have focused on classifying opinions based on their positive or negative polarity. However, there have been some quantitative studies on multi-class opinion classification. Sentiment analysis involves several stages, including data preprocessing [2], rule-based techniques, feature extraction [3], polarity classification, and opinion summarization [4]. Initial approaches in this field relied on machine learning methods, but today, linguistic rules and NLP algorithms are used to enhance the accuracy and efficiency of these approaches.

The Persian language has unique characteristics that challenge NLP tasks, such as complex morphology and syntax. Research on sentiment analysis in the Persian language can improve NLP models for this language, which can be used in other NLP applications. One of the most critical issues, particularly in social networks, is the scarcity of Persian language data and the limited research in opinion mining, especially in aspect detection in the Persian language using semantic tools.

Although document-level or sentence-level text classification can be useful in many cases, it does not provide the necessary detail for many other applications. A document with a positive opinion about a particular entity does not mean that the author has positive opinions about all aspects of the entity. Similarly, a document with a negative opinion does not mean that the author hates everything. In a typical opinion document, the author writes both positive and negative aspects of the entity, although public sentiment about the entity may be positive or negative. Sentiment classification at the document and sentence level does not provide such information, so to obtain these details, classification at the aspect level must be performed. This research focuses on improving aspect-based sentiment analysis (ABSA) on Persian language datasets. For example, the dataset under investigation includes user opinions about the purchased food items. By using ABSA, various aspects of the opinions can be examined. The necessity of this research becomes apparent when precise retrieval of information from a text or opinion requires knowledge of its sentiments towards specific topics or aspects. Complementary tools such as WordNet or ontologies will be used to enhance the accuracy of search and retrieval of words, which can assign multiple meanings to aspects in the analyzed text. This gap existed in previous studies where aspects were only searched with their exact word in the text. Furthermore, in the Persian language, there are figurative and semantic ambiguities hidden behind words that previous methods lacked sufficient capabilities to improve due to the absence of additional tools and information like dictionaries and WordNet. Therefore, it is expected that more accurate results can be achieved by employing such tools. Additionally, this research demonstrates the potential use of a pre-trained BERT model [5] and the utilization of input from both the target sentence and the aspect in ABSA.

The research methodology used for this study specifically employed an experimental research design with a pre-test/post-test approach. In this method, we started with an existing sentiment analysis model in ABSA. We introduced changes to the model by adding synonyms and meanings to the dataset, particularly in aspects. These changes were considered experimental modifications. We then evaluated the performance of the modified model by comparing its accuracy and F1 score with the original model or other advanced models on an evaluation dataset. This forms the pre-test/post-test design.

Main contributions of our work are as follows:
- Fine-tuning the ParsBERT model.
- Increasing accuracy and F1 score through enriching aspects.
- Collecting a set of synonymous words like a lexicon.

## 2. Related Works

Sentiment analysis is currently a field at the intersection of NLP and information retrieval (IR), with shared aspects of information extraction and text mining [6]. Sentiment analysis methods can be categorized into machine learning, hybrid, and lexicon-based approaches. In the machine learning approach, a set of feature vectors is selected, and a labeled corpus is used to train a classifier that can be applied to unlabeled text. Feature selection plays a crucial role in the success of classification. Different types of single words or multi-word expressions are typically chosen as feature vectors. Other proposed features include the number of positive words, the number of negative words, and the document's length. Support Vector Machines (SVM) and Naive Bayes algorithms are the most commonly used classification techniques [7]. Joachim suggested using SVM for text classification tasks to leverage its stability in high-dimensional spaces [8]. Naive Bayes classifier performs well in classifying large datasets. The problem with this approach is that it assumes independence among all features, so the Naive Bayes classifier achieves perfect accuracy only if the features are independent. Adam *et al*. used Naive Bayes for Sentiment analysis on movie reviews [9].

The lexicon-based approach typically relies on pre-annotated dictionaries or word collections. Each word in a text is compared to the lexicon, and if a word exists in the lexicon, its sentiment polarity value is added to the overall sentiment score. An example of the application of lexicon-based methods is SentiWordNet as an improved lexical resource for sentiment analysis and opinion mining, as presented by Baccianella et al. [10].

In the mentioned approaches, sentiment analysis is performed on the entire document based on the polarity of words, categorizing it as positive or negative. The text is not classified based on the sentiment scores of specific words. Approaches based on ontology also perform a similar task using Ontology Web Language (OWL) and Resource Description Framework (RDF) languages. In this approach, an ontology is created, opinions are extracted, the extracted opinions are labeled using the ontology, sentiment scores are assigned by creating a lexicon representing each expression's sentiment polarity, and an overall sentiment score is calculated [11]. Thakor and S. Sasi [12] introduced an ontology-based sentiment analysis process for social media content.

In 2016, Schouten *et al*. conducted a study on ABSA [13]. ABSA focuses not only on finding overall sentiments related to an entity but also on finding sentiments for its specific aspects. Sun *et al*. [14] demonstrates that because the input in the BERT model can be either a single sentence or a sentence pair, the task of ABSA can be transformed into a sentence pair classification approach. In this method, the first sentence represents a text containing sentiments towards one or multiple aspects, and the second sentence (auxiliary sentence) represents a specific aspect of interest. The pre-trained BERT model takes this sentence pair as input and performs a sentence pair classification. The results show that BERT can be considered one of the best tools for ABSA. To enhance the capability of aspect-based polarity classification, embedding of aspects and context from one side [15] and/or hierarchical attention-based method from the other side can play a significant role [16]. Clarifying the role of these elements further emphasizes their contribution to the success of BERT in ABSA. Embeddings, serving as contextualized representations of words and aspects, enable BERT to capture nuanced relationships within sentences. Context, achieved through the utilization of segment embeddings and positional embeddings, ensures a holistic understanding of the interplay between aspects and their surroundings. The hierarchical attention-based method allows BERT to adaptively focus on crucial aspects within a sentence, enhancing the model's ability to discern subtle sentiment nuances. By comprehensively elucidating the roles of embeddings, context, and hierarchical attention-based methods, we provide a deeper insight into the mechanisms that propel BERT as an effective tool for aspect-based sentiment analysis.

Farahani *et al*. [17] introduce a single-language Persian language model called ParsBERT, based on the BERT architecture, which is pre-trained and fine-tuned on a large Persian language dataset. They utilized the BERT architecture to build a pre-trained language model (PLM) for Persian. The first aspect-level dataset for the Persian language, named SentiPers, was introduced in [18], which included over 26,000 sentences with 21,375 aspects. Dastgheib *et al*. [19] used the SentiPers dataset at the sentence level for text sentiment classification. This reference proposed a combined method of structured correspondence learning (SCL) and convolutional neural network (CNN). The results showed that the proposed method outperforms a simple CNN and a naive Bayes classifier.

PourMostafa Roshan Sharami and Sarabestani [20] also utilized the SentiPers dataset at the sentence level for sentiment analysis. Bidirectional LSTM and CNN were two deep learning architectures examined in this research. Their results demonstrated that both methods perform better than baseline models, with Bidirectional LSTM achieving the best results. Ataei *et al*. [21] presented a manually annotated Persian aspect-based sentiment analysis dataset called Pars-ABSA, which consists of 5,602 unique opinions and 10,002 aspects. A detailed comparison of the mentioned methods is provided in Table 1.

Most of the research and work conducted in opinion mining and sentiment analysis has been done on non-Persian languages, mainly English. However, recently in Iran, the potential for sentiment analysis and understanding people's beliefs in the community has emerged due to increasing public attention to web pages and social networks. People refer to web pages or social networks for daily activities such as purchasing items or watching movies to gain better insights and make informed decisions. Essentially, analyzing user reactions and sentiments towards a topic by other users increases societal awareness regarding the content published in the virtual space. The lack of algorithms, their complements, and datasets in the Persian language has led to establishing the foundations for this research.

This study expands upon the aforementioned resources by improving them in various ways. Firstly, additional tools such as lexicons are utilized to enhance the efficiency and accuracy of the language model. Secondly, the study incorporates the latest sentiment analysis algorithm, BERT, and a separate dataset to improve the accuracy of the BERT model. Lastly, a different algorithm for ABSA is employed, along with a complementary tool such as a lexicon, to further enhance the accuracy of the BERT model. With the implementation of these advancements, this study builds upon previous research foundations and takes significant steps in sentiment analysis.

## 3. Proposed Approach

In this section, first, the idea of the proposed approach is presented, followed by a description of the main components of the proposed method. Most of the research and work conducted in the field of opinion mining and sentiment analysis has been carried out on non-Persian language datasets, particularly English.

**Table 1. Comparison of sentiment analysis methods.**

| Strengths | Weaknesses | Performance | Method | References |
|---|---|---|---|---|
| • Simpler interpretation compared to other methods<br>• Quick and easy computation<br>• No impact from irrelevant features | • Assumption of feature independence | • Estimating the probability of positive or negative occurrence of data | Naive Bayes | [22], [23] |
| • Can be used for data with irregular and unknown distributions<br>• Stable in high-dimensional spaces<br>• Usually not prone to overfitting<br>• Performs well when there is a clear boundary between classes | • Long training time for large datasets<br>• Difficulty in understanding and interpreting the final model<br>• Challenging fine-tuning of hyper parameters<br>• Dependency on the domain | • Creating word clouds to distinguish positive and negative classes | Support Vector Machines (SVM) | [24], [25] |
| • Does not rely on statistical dependencies for random variables<br>• Obtaining appropriate distribution | • Requirement of more resources and manual interpretations | • Utilizing data from previous experiments to find the best probability distribution | Maximum Entropy | [26] |
| • Does not require training because the polarities are pre-defined and given as input to the algorithm | • Ambiguity in words with multiple meanings | • Using predefined lexicons of words and associated sentiments for text analysis | Dictionary-based | [27], [28] |
| • Similar to the dictionary-based method, it does not require training | • Difficulties in creating a specialized glossary for a specific topic | • Utilizing a large dataset for training a machine learning model to identify sentiments in new texts<br>• Analyzing patterns of words and phrases in the corpus with positive, negative or neutral sentiments | Corpus-based | [29] , [30] |
| • Inherits stability from linguistic methods and high accuracy from machine learning. | • Can be more complex and require more data and advanced algorithms.<br>• Susceptibility to overfitting, leading to inaccurate sentiment analysis in real-world scenarios. | • Using a combination of approaches for text analysis, such as:<br>• Identifying sentiment-bearing words from the lexicon.<br>• • Analyzing patterns in a large collection of texts.<br>• • Training machine learning models to detect sentiments in new texts | Mixed | [31], [32] |

Recently, in Iran as well, due to the increasing attention of people to web pages and social networks, the potential for analyzing sentiments and opinions of society members has emerged. This is because people refer to web pages or social networks for their daily tasks such as buying goodsor watching a movie to gain better insights for similar purchases or activities.

Essentially, analyzing users' reactions and sentiments towards a piece of content by other users leads to an increased awareness of the community members regarding the published content in the virtual space. The lack of algorithms, their complements, and datasets in the Persian language has prompted the foundation of this research.

The model used in this study is based on the BERT model architecture, which consists of a multi-layer bidirectional transformer. Specifically, this model utilizes the original BERT base configuration, which includes 12 hidden layers, 12 attention heads, and a hidden size of 768. The total number of parameters in this configuration is 110 million. This configuration was chosen because our task is not extensive enough to warrant the use of Large BERT, ensuring a balanced utilization of resources while maintaining optimal performance for our specific task. According to the original purpose of BERT pre-training, it serves two tasks:

- Masked language modeling: To train the model, a masked language model (MLM) is used to predict randomly masked tokens using cross-entropy loss. For this purpose, 15% of the tokens are randomly selected from the input sequence. Among these selected tokens, 80% are replaced with an exclusive [MASK] token, 10% are replaced with a random token, and 10% remain unchanged.
- Next sentence prediction (NSP): The model learns to predict whether the second sentence in a sentence pair is the actual subsequent sentence of the first sentence or not. In the original BERT paper, it is argued that removing NSP task prior to training can reduce the model's performance in certain tasks. Therefore, they included NSP in their model to ensure high performance across various tasks.

For model optimization, the AdamW optimizer [33] is used $\beta_1 = 0.9$ and $\beta_2 = 0.98$ for 1 million steps. The batch size is set to 32, and each sequence has a maximum length of 512 tokens. Finally, the learning rate is set to 0.0001. The Exponential Moving Average optimizer calculates the moving average of gradients and squared gradients for each parameter, facilitating parameter updates during training. It incorporates hyperparameters for decay rates and initial learning rates, allowing customization for task-specific optimization. In the context of the BERT language model, the Adam optimizer is employed to minimize the loss function during training. This loss function measures the disparity between predicted outputs and actual labels, enabling BERT to learn precise predictions across diverse natural language processing tasks.

The sub-word tokenization, which is essential for improved performance, is obtained using the WordPiece method. WordPiece acts as an intermediary between byte-pair encoding and single-word language modeling approaches. The vocabulary consists of 100,000 tokens, including specific tokens for BERT, such as [PAD], [UNK], [CLS], [MASK], [SEP], and [##], which are used as prefixes for tokenization to establish word relationships. Sub-word tokenization is essential for improved performance as it allows the model to effectively handle complex word structures and variations, capturing fine-grained details in language. This approach enhances the model's ability to understand and represent the morphological intricacies of words, contributing to its overall efficiency in language understanding and downstream tasks.

Figure 1 illustrates the overall structure of the proposed method. The BERT sentence-embedding layer takes the input sentence and calculates token-level representations using the contextual information of the entire sentence. Since we intend to use BERT as a text classifier, a unique linear layer is added on top of it. As the input data is fed, the entire model is fine-tuned, with the pre-trained BERT and the additional non-pre-trained classification layer, specifically tailored for the aspect, to achieve precise adjustments.

The BERT model takes a two-sentence input, consisting of a review sentence and an aspect phrase, and predicts the sentiment polarity towards the aspect phrase as the output. In other words, the review is provided as the first sentence of the model, while the aspect phrase serves as a supporting sentence, filling the second position. Using a supporting sentence (the second sentence) enhances the performance of the BERT model for ABSA tasks. Before feeding the inputs into the BERT model, we enrich each aspect using the employed lexicon [34]. The lexicon used in this study contains synonyms of one or two words, encompassing 19,899 different terms. The synonym words are extracted from lexicons such as Dehkhoda and Moin, which mostly include only two-part words. This lexicon has been provided by the Abadis website. Incorporating a semantically relevant lexicon is pivotal in elevating BERT's effectiveness for semantic analysis and opinion modeling. The addition of this lexicon contributes significantly to BERT's capacity to discern nuances in language and context. BERT, while powerful, may encounter challenges in capturing the intricacies of sentiment, especially in languages with rich contextual variations like Persian. The lexicon serves as a valuable supplement, providing additional semantic cues and context-specific information.

By integrating a lexicon tailored to the Persian language, our approach augments BERT's semantic capabilities, enabling a more nuanced understanding of sentiment expressed in user-generated content. The synergistic combination of BERT and the lexicon facilitates a comprehensive analysis of sentiments associated with various aspects. This integration is particularly beneficial in capturing subtle sentiment variations, which may be crucial in domains where context plays a pivotal role, such as understanding user opinions on a diverse e-commerce platform like 'Digikala.'

The enhanced functionality achieved through this integration is instrumental in elevating the overall performance of the sentiment analysis algorithm. It empowers the model to not only recognize sentiment polarity but also grasp the underlying semantic nuances, leading to more accurate and context-aware predictions. Our experimental results showcase the tangible impact of this integration, demonstrating improved accuracy and F1 scores, validating the promising potential of this approach in advancing the capabilities of deep neural modules for sentiment analysis.

As shown in Figure 1, the appropriate tokenization is first selected using the proposed model. In the second stage, which is the data preparation stage, the length of the reviews is calculated in terms of the number of words, and any reviews with unusually long lengths are removed. In the next step, the second input, which is the aspect, is enriched using the proposed lexicon, and the input pairs are tokenized. The [CLS] token is added to the beginning of the review or the first sentence. After the completion of the review, the [SEP] token is appended to continue the assumed string. This separates the review sentence and the aspect phrase using the [SEP] token, and then tokenization takes place. Then an attention layer is generated, which, along with the tokenized inputs from the previous step, is converted into a tensor. The next part involves three embedding stages: 1-token embedding 2-position embedding 3-segment embedding. In the token embedding stage, the WordPiece method is employed. In the position embedding stage, the positions of the words in the respective sentence are encoded into a vector. All three embeddings are summed together with the elements they generated, resulting in a representation of dimensions (1, n, 768), which serves as the input to the BERT encoding layer. The next part is the transformer layers, which are a type of neural network architecture designed for processing sequential data, such as natural language texts. Each transformer layer in BERT consists of two sub-layers: a multi-head self-attention layer and a feed-forward neural network layer.

The final step is passing the combined representation through a classification layer, which is a simple feed-forward neural network that assigns a sentiment label to the sentence-level representation. For binary sentiment analysis, the classification layer may produce a single scalar value indicating the probability of the input text being positive or negative. For multi-class sentiment analysis tasks like this study, the classification layer may output a probability distribution over multiple sentiment labels.

In this study, the specific language model used is called "bert-fa-base-uncased-clf-digimag" [17]. The model was pre-trained on a large corpus of Persian news articles using a MLM objective and then fine-tuned on a labeled dataset of news articles. In this study, fine-tuning was performed with careful consideration of the specific dataset used.

In the fine-tuning process, particular considerations were made to optimize the model for our dataset. The AdamW optimizer was utilized with specific hyper-parameters, including $\beta1 = 0.9$, $\beta2 = 0.98$, and a learning rate of $0.0001$ over 1 million steps. The batch size, sequence length, and sub-word tokenization using the WordPiece method were carefully chosen to tailor the model to the characteristics of our dataset, aiming for optimal performance and decreasing the complexity of the model.

## 4. Experimental Results
### 4.1. Dataset
In this section, we analyze the used data statistically. The ParsiNLU dataset [36] in ABSA task consists of 500 customer reviews about purchasing food products on the Digikala website in the Persian language. Each review has been annotated for ABSA. The format of this dataset is XML.

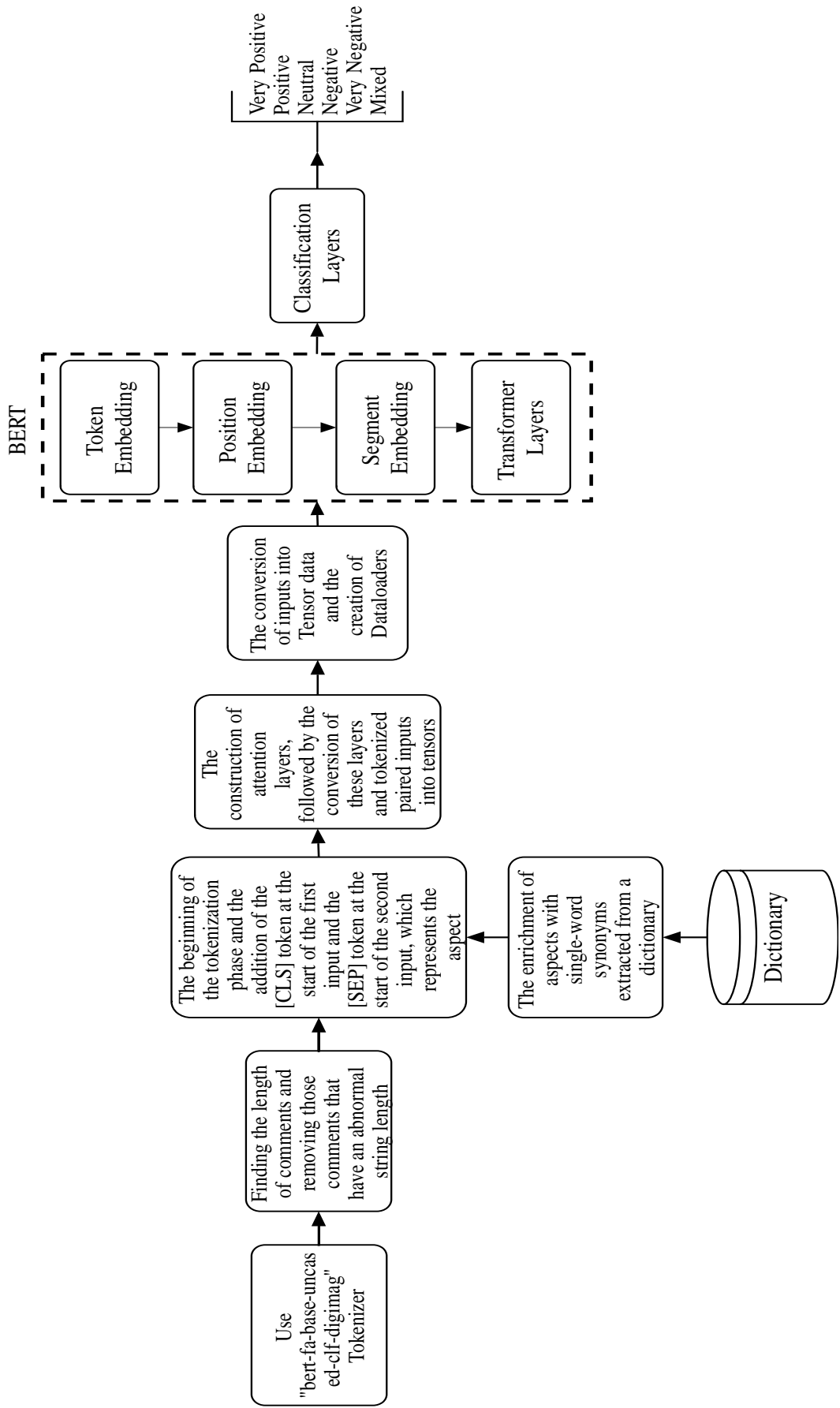

**Figure 1. The overall structure of the proposed method and the interplay between its components.**

The most important aspect of the used dataset for NLP models is its domain, which, as mentioned, is related to food and edible products. The dataset annotation was done manually by a team of experts following the guidelines for ABSA. Each aspect may have one of the following sentiment labels in the label field: very negative (-2), negative (-1), neutral (0), positive (1), very positive (2), mixed (3), and for those aspects that are not mentioned in the review text, the label -3 is considered. The statistical description of the assumed dataset is presented in the following figures. Figure 2 illustrates the distribution of the analyzed categories in the dataset.

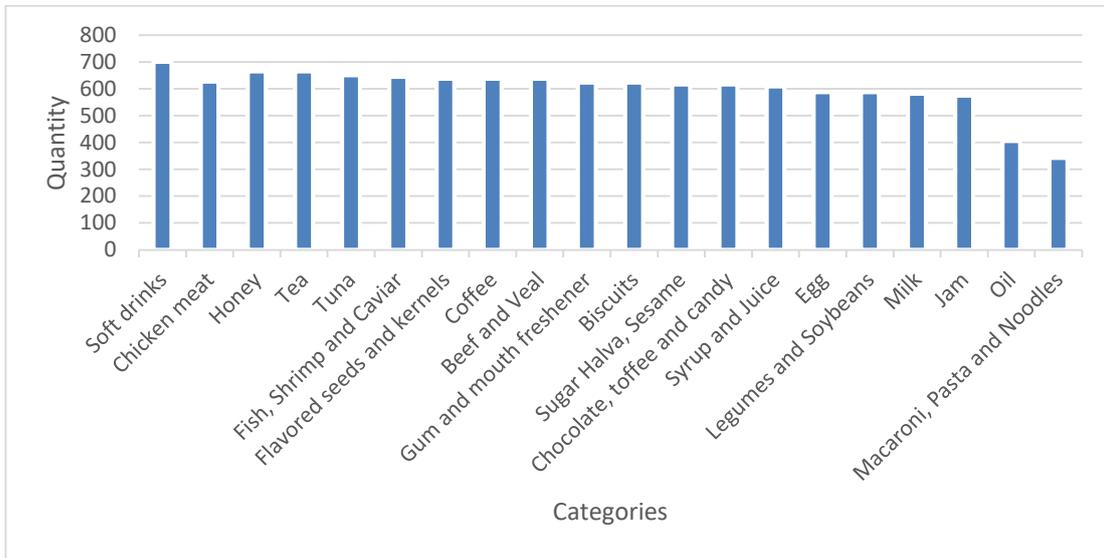

**Figure 2. Distribution of opinion categorizations by food type.**

Figure 3 represents the distribution of labels in the dataset, which have been annotated by several users. As observed in the figure the distribution of the labels of the aspects is imbalanced, with the majority of opinions (69.5%) not leaving comments on specific aspects. However, 8% of the opinions were positive, 6.8% were negative, 6.6% were very negative, 6% were very positive, 0.3% were neutral, and 1.8% were mixed. The reason for the highest share of "No comment for the aspect" is that not all aspects are mentioned in the text of the opinions. This imbalanced distribution of labels poses a challenge, potentially impacting overfitting during model training, and efforts have been made to address this issue by considering strategies such as adjusting class weights or employing techniques to handle imbalanced datasets effectively.

### 4.2. Evaluation metrics & experiments
This research work uses precision, recall, and F1 score as evaluation metrics to compare the performance of the proposed method with the models reported in the literature.

For the implementation of this research, Google Colab was used, which utilizes the Python language. Due to the use of tensors, we needed to use GPUs. NVIDIA has provided a specialized graphics processor for tensor processing called V100, which Google Colab utilizes.

To compare the results, the evaluation table from reference [35] was used, where accuracy is the metric of comparison in that table. mBERT is a PLM developed by Google to understand text in multiple languages. It is based on the Transformer architecture and has achieved advanced performance in various NLP tasks. The next model examined was wikiBERT, developed by researchers at the University of Los Angeles, designed to improve the performance of NLP tasks such as text classification, question answering, and language translation. mT5, developed by Google, is a PLM designed for a diverse range of NLP tasks, including machine translation, summarization, and question answering. It is based on the Transformer architecture and trained on a large corpus of texts from multiple languages. Unlike other multilingual models that focus on understanding text in multiple languages, mT5 is specifically designed for generating text in multiple languages. The main difference between this model and mBERT lies in their primary development objectives and the trained datasets.

As seen in Table 2, the highest accuracy value belongs to the mT5 model, which is 58.9, while the enriched model in this research achieved an F1 score of 61.7 and a computed accuracy of 88.2. The reason for this improvement lies in two aspects emphasized in this research. Firstly, the selected PLM is suitable for this dataset, and secondly, with the help of the prepared lexicon, we were able to enhance the model's performance and increase the accuracy from 53.9 to 88.2, and the F1 score from 58.3 to 61.7. This improvement can occur for three reasons. From the perspective of increased precision, by combining synonyms, the model can better identify and extract all references to an aspect in the text, even if different words or phrases are used. This can contribute to improved sentiment analysis accuracy by ensuring that all relevant information is preserved. From the perspective of increased coverage, by adding the meaning of each aspect, the model can better understand the context in which each aspect is mentioned and improve the coverage of analysis. This can help ensure that all relevant aspects are included and that the associated sentiments for each aspect are correctly identified. The third reason is that the selected model has been pre-trained on a corpus more similar to the dataset of this research, and by adding meaning to the aspect domain, we were able to surpass the performance of the latest ParsBERT language model.

**Table 2.** Results obtained on the dataset for the sentiment analysis task in the aspect-specific sentiment.

| Setup | Model ↓ -Task→ | Performance |
|---|---|---|
| | mBERT (base) | 53.9 |
| | WikiBERT (base) | 56.5 |
| | ParsBERT (base) | 53.9 |
| Trained on persiNLU | mT5 (small) | 52.4 |
| | mT5 (base) | 52.9 |
| | mT5 (large) | 58.1 |
| | mT5 (XL) | 58.9 |
| | **Proposed model (lexicon-based)** | **88.2** |

To demonstrate the improvements achieved by our approach, we constructed a structured evaluation framework. This framework consisted of a table featuring a range of real-world reviews in Persian, each accompanied by corresponding aspects, actual labels, labels predicted by the proposed model, and labels predicted by the based model.

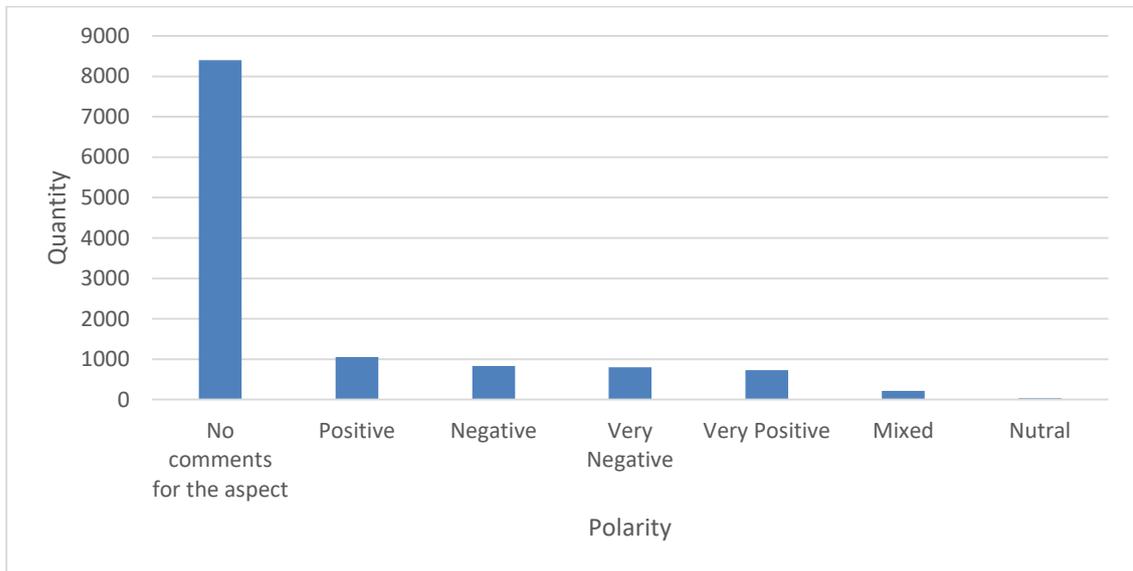

**Figure 3.** Distribution of labels.

Table 3 representation allowed us to showcase the performance disparity between the two models across various instances. Notably, our proposed model consistently demonstrated superior predictive capabilities, aligning more accurately with the actual sentiments expressed in the reviews, especially in cases where synonymous expressions were involved. By doing so, the model can detect synonyms of aspects in some comments that do not include previous aspect words, showcasing its advanced semantic understanding and improved performance in capturing subtle variations in sentiment expression. As it can be seen from the table, in the first review, the ParsBERT model detected 'very positive' for 'Taste' aspect, while the actual label is 'Neutral,' and our proposed model predicts 'Neutral,' aligning with the actual label. In the second review, the proposed model predicts Neutral for the 'worth buying' aspect, consistent with the actual label, whereas the ParsBERT model predicts 'Negative.'

**Table 3. Performance comparison of proposed and based models for sentiment analysis on enriched PersiNLU dataset.**

| Review | Aspect term | Corresponding aspect terms | Actual label | Proposed model | Based model |
|--------|-------------|----------------------------|--------------|----------------|-------------|
| Persian:<br>خیلی به بوی گوشت حساس خیلی خوب بود چربی اندازه و تازه گوشت صورتی بود ممنون از گوشت خوب باز میخرم.<br>English:<br>I am very sensitive to the smell of meat. It was very good. The fat was ok and the meat was pink. Thank you. I will buy it again. | Persian:<br>طعم<br><br>English:<br>Taste | Persian:<br>طعم، چاشنی، مزه، چشایی، ذائقه و مذاق<br>English:<br>Taste | Neutral | Neutral | Very Positive |
| Persian:<br>کیفیتش بد بود و درکل بوی ماهی کیلکا رو گرفته بودند.<br>English:<br>The quality was bad and they had taken the smell of Kilka fish out of it. | Persian:<br>ارزش خرید<br><br>English:<br>Worth buying | Persian:<br>ارزش خرید، ارزشمند، شایسته خریدن و لایق خریدن<br>English:<br>Worth buying | Neutral | Neutral | Negative |
| Persian:<br>در مقایسه با سایر برندهای موجود در بازار با توجه به حراجی که داشت ارزانتر بود.<br>English:<br>Compared to other brands in the market, it was cheaper due to the discount. | Persian:<br>کلی<br><br>English:<br>General | Persian:<br>کلی، عام، عمومی، همگانی، در مجموع، فراگیر، جامع و در کل<br>English:<br>General | Positive | Positive | Negative |
| Persian:<br>درمقایسه با سایر برندهای موجود در بازار کیفیتش قابل قبول بود<br>English:<br>Compared to other brands in the market, its quality was acceptable. | Persian:<br>ارزش خرید<br><br>English:<br>Worth buying | Persian:<br>ارزش خرید، ارزشمند، شایسته خریدن و لایق خریدن<br>English:<br>Worth buying | Positive | Positive | Negative |

To prove the performance improvement compared to the latest developed language model, ParsBERT, we will use the precision-recall curve, confusion matrix, accuracy, recall, and F1 score for each class. Figure 4 shows the overall performance metric of the model for correctly classifying positive samples while minimizing false positives. It considers both precision and recall, which are important for evaluating a model's performance on imbalanced datasets. As observed, the proposed enriched model in this study achieves a higher level, indicating the superiority of the proposed enriched model with semantics over the state-of-the-art ParsBERT language model.

By comparing the Confusion Matrix in Figure 5 for the proposed model and Figure 6 for the ParsBERT model, we can also highlight the superiority of the proposed model in terms of having a higher number of correct predictions.

For calculating the three metrics of accuracy, recall, and F1, the sklearn library was used, and the results are presented in Tables 4 and 5, providing a breakdown for each of the six classification classes.

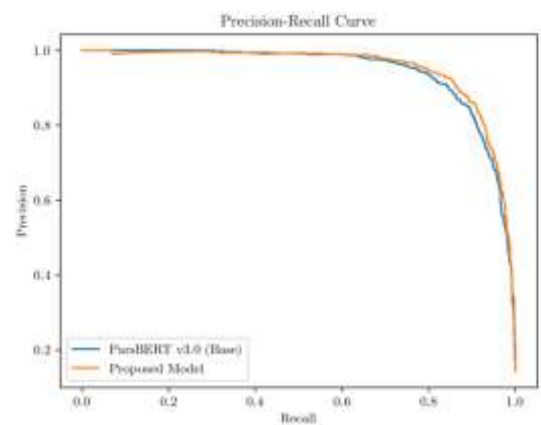

**Figure 4. Comparison of area under the precision-recall curve.**

By comparing the numbers of these three metrics in the tables, the superiority of the proposed model can be observed.

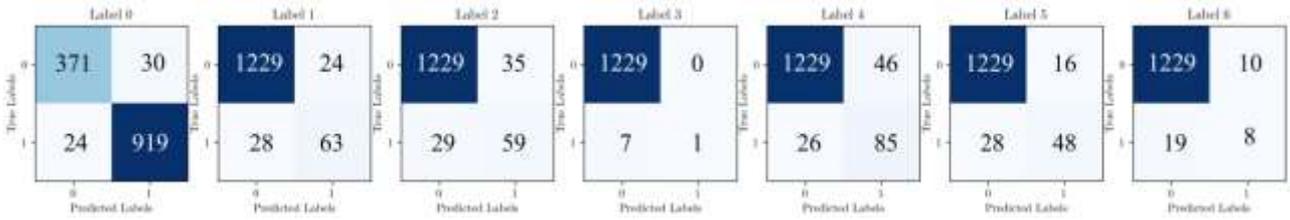

**Figure 5.** Confusion matrix of the proposed model.

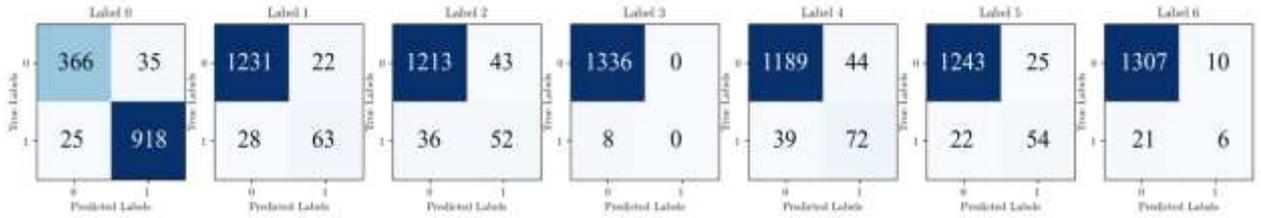

**Figure 6. Confusion matrix of the ParsBERT model.**

**Table 4. Evaluation metrics of the proposed model.**

|  | Precision | Recall | F1-Score | Support |
|---|---|---|---|---|
| 0 | 0.97 | 0.97 | 0.97 | 943 |
| 1 | 0.72 | 0.69 | 0.71 | 91 |
| 2 | 0.63 | 0.67 | 0.65 | 88 |
| 3 | 1.00 | 0.12 | 0.22 | 8 |
| 4 | 0.65 | 0.77 | 0.70 | 111 |
| 5 | 0.75 | 0.63 | 0.69 | 76 |
| 6 | 0.44 | 0.30 | 0.36 | 27 |
| Accuracy |  |  | 0.88 | 1344 |
| Macro Avg. | 0.74 | 0.59 | 0.61 | 1344 |
| Weighted Avg. | 0.88 | 0.88 | 0.88 | 1344 |

**Table 5. Evaluation metrics of the ParsBERT model.**

|  | Precision | Recall | F1-Score | Support |
|---|---|---|---|---|
| 0 | 0.96 | 0.97 | 0.97 | 943 |
| 1 | 0.74 | 0.69 | 0.72 | 91 |
| 2 | 0.55 | 0.59 | 0.57 | 88 |
| 3 | 0.00 | 0.00 | 0.00 | 8 |
| 4 | 0.62 | 0.65 | 0.63 | 111 |
| 5 | 0.68 | 0.71 | 0.70 | 76 |
| 6 | 0.38 | 0.22 | 0.28 | 27 |
| Accuracy |  |  | 0.87 | 1344 |
| Macro Avg. | 0.56 | 0.55 | 0.55 | 1344 |
| Weighted Avg. | 0.86 | 0.87 | 0.86 | 1344 |

Visualizing accuracy and loss of training and testing data is an important aspect of evaluating the performance of sentiment analysis models. By plotting these metrics, we can assess how well the model performs on both the training and testing data. Ideally, we would like to see a decreasing trend in the loss function and an increasing trend in accuracy. Figures 7 and 8 depict these trends visually.

Shifting focus to the proposed model, intentional efforts were made to mitigate potential overfitting by reducing complexity and fine-tuning hyperparameters. Importantly, the model hasn't seen the whole dataset during training, ensuring its ability to generalize to unseen instances. While no specific action was taken to address dataset imbalance, the model's performance is evaluated with awareness of the data distribution. Ongoing efforts explore strategies for handling imbalanced data, aiming to enhance the model's generalization to new, unseen instances. It's worth noting that we are unable to balance the dataset due to the significant loss of data, emphasizing the importance of optimizing model performance within the constraints of available data resources.

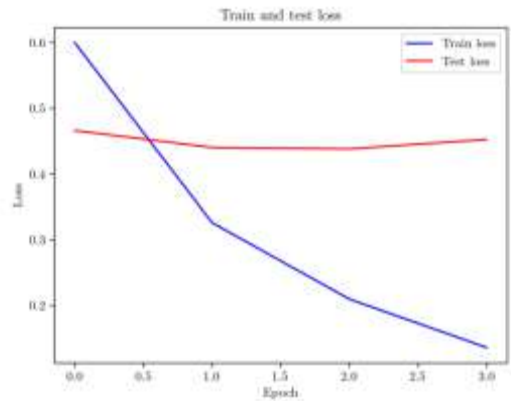

**Figure 7. Learning loss plot.**

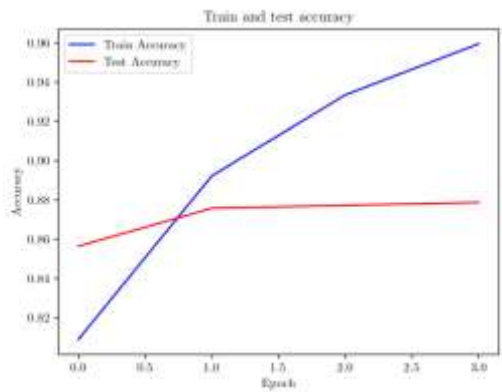

**Figure 8. Learning accuracy plot.**

## 5. Concluding Remarks

In this paper, it was shown that the integration of synonyms to the aspect terms can improve the performance of ParsBERT language model. The prominent results highlight the role of semantic enrichment in improving the accuracy of sentiment classification and emphasize the potential for further research and development in ABSA techniques. By combining a richer understanding of language related to aspects, the modified model demonstrates enhanced capabilities in capturing subtle emotional differences.

Our approach introduces a novel dimension to sentiment analysis by integrating synonyms and semantic enrichment, setting it apart from existing methods. Unlike traditional methods that may struggle with synonymous expressions or lack the depth of semantic understanding, our model excels in capturing nuanced sentiments through the identification of synonyms for aspects. This innovative integration enhances the model's performance, allowing it to detect and interpret synonymous expressions in reviews accurately. Our research uniquely contributes to the field by bridging the gap in semantic analysis, thereby offering an advanced and more nuanced perspective in sentiment classification compared to conventional methods.

The improved accuracy and reduced losses observed in the modified model underscore the value of incorporating semantic information in ABSA tasks. This progress opens doors to more precise sentiment analysis, enabling businesses and organizations to extract valuable insights from customer feedback, make informed decisions, and foster improved customer satisfaction, loyalty, and engagement. The enhanced precision in sentiment classification directly contributes to more informed decision-making processes and effective customer relationship management. Businesses can better understand customer sentiments, preferences, and feedback, tailoring their products, services, and strategies to meet customer expectations. The model's improved performance contributes to more accurate sentiment-driven insights, empowering organizations to make data-driven decisions, refine marketing strategies, and proactively address potential issues. This holistic improvement ultimately enhances overall business intelligence and operational efficiency, showcasing the significant implications of incorporating semantic enrichment techniques in sentiment analysis for both ABSA tasks and broader applications in natural language processing.

Future research directions may explore additional semantic techniques such as word embedding or suitable text representations to further enhance the performance of sentiment analysis in ABSA. Examining domain-specific adaptations can also be beneficial as different industries and domains may have unique aspects and expressions of sentiment. Additionally, the integration of advanced deep learning models such as recurrent neural networks or transformers holds promise for achieving higher accuracy and stability in ABSA.

Overall, this research paves the way for more sophisticated and dynamic sentiment analysis applications in various domains by improving sentiment analysis models through semantic enrichment. The results serve as a foundation for future research and support the advancement of sentiment analysis methods and their practical implementations in real-world scenarios.

# ارتقاء تحلیل احساس منظر–پی با استفاده از پارس‌برت در زبان فارسی

فرید آریایی۱، مریم طایفه محمودی۲،* و علی معینی۳

۱ گروه مهندسی کامپیوتر، دانشگاه تهران، پردیس بین المللی کیش ، کیش، ایران.

۲ گروه سامانه‌های پردازش و تحلیل داده‌ها، پژوهشکده فناوری اطلاعات، پژوهشگاه ارتباطات و فناوری اطلاعات، تهران، ایران.

۳ دانشکده علوم مهندسی، دانشکدگان فنی، دانشگاه تهران، ایران.

## چکیده:

در عصـر اسـتفاده گسـترده از اینترنت و حاکمیت شـبکه‌های اجتماعی، پژوهشـگران با چالش‌های قابل توجهی در متن کاوی فارسـی، از جمله کمبود مجموعه داده‌های کافی به زبان فارسـی و ناکارآمدی مدل‌های زبانی موجود، روبرو هسـتند. مقاله حاضـر این چالش‌ها را به ویژه با هدف افزایش کارایی مدل‌های زبانی سازگار با زبان فارسی مورد توجه قرار می‌دهد. با تمرکز بر افزایش اثربخشی تحلیل احساسات، رویکرد ما از نوعی روش‌شناسی منظر–پی با به کار بردن مدل پارس‌برت و فرهنگ لغات مرتبط با موضـوع اسـتفاده می‌کند. این مطالعه بر تحلیل احسـاسـات نظرات مسـتخرج از کاربران وبگاه فارسـی زبان 'دیجی‌کالا' متمرکز اسـت. نتایج تجربی نه تنها توانایی‌های معنایی برتر روش پیشـنـهادی را برجسـته می‌کند، بلکه افزایش کارایی آن را با دقت ۸۸٫۲٪ و امتیاز F1 ۶۱٫۷ نشـان می‌دهد. اهمیت ارتقاء مدل‌های زبانی در این زمینه در نقش حیاتی آن‌ها در اسـتخراج احسـاسـات دقیقی نهفته اسـت که محتوای آن را کاربران تولید می‌کنند و د ستآخر با افزایش کارایی و دقت، به پیـشرفت حوزه تحلیل احـساسات در متن کاوی به زبان فارسی می‌انجامد.

**کلمات کلیدی:** نظرکاوی، تحلیل احساس، تحلیل احساس منظر–پی، ابهام زدایی معنایی واژگان، وردنت